\pdfoutput=1

\documentclass[11pt]{article}

\usepackage{EACL2023}

\usepackage{times}
\usepackage{latexsym}
\usepackage{adjustbox}
\usepackage[T1]{fontenc}

\usepackage[utf8]{inputenc}

\usepackage{microtype}

\usepackage{inconsolata}

\title{Fairness in Language Models Beyond English: Gaps and Challenges}

\author{Krithika Ramesh, Sunayana Sitaram, Monojit Choudhury\\
  Microsoft Corporation\\
  \texttt{\{t-kriramesh, sunayana.sitaram, monojitc\}@microsoft.com}}

\begin{document}
\maketitle
\begin{abstract}

With language models becoming increasingly ubiquitous, it has become essential to address their inequitable treatment of diverse demographic groups and factors. Most research on evaluating and mitigating fairness harms has been concentrated on English, while multilingual models and non-English languages have received comparatively little attention. This paper presents a survey of fairness in multilingual and non-English contexts, highlighting the shortcomings of current research and the difficulties faced by methods designed for English. We contend that the multitude of diverse cultures and languages across the world makes it infeasible to achieve comprehensive coverage in terms of constructing fairness datasets. Thus, the measurement and mitigation of biases must evolve beyond the current dataset-driven practices that are narrowly focused on specific dimensions and types of biases and, therefore, impossible to scale across languages and cultures.

\end{abstract}

\section{Introduction}
    Language models are known to be susceptible to developing spurious correlations and encoding biases that have potentially harmful consequences in downstream tasks. Whilst prior work has documented these harms \cite{dev-etal-2021-harms} \cite{10.1145/3442188.3445922} \cite{https://doi.org/10.48550/arxiv.2210.07700}, there remains much to be studied and criticism for the existing research (or lack thereof) that remains to be addressed.
    
    In the context of language models, fairness can manifest in two forms; \textbf{\textit{representational}} and \textbf{\textit{allocational}} harms. \textbf{Representational} harms generally refer to cases where demographic groups end up being misrepresented. This includes stereotypes and negative associations with these groups and even a lack of acknowledgment of certain groups that are underrepresented in the data. \textbf{Allocational} harms, on the other hand, refer to the inequitable distribution of resources and opportunities to groups with different demographic attributes associated with them. The nature of allocational harms can vary based on the \textbf{sociocultural, economic, and legal} settings where the system has been deployed. However, it can also take shape in terms of the model's functionality across languages with fewer resources \cite{Choudhury_Deshpande_2021, liu-etal-2021-visually}. While current literature adopts a Euro-American-centric view of fairness, work such as \citet{DBLP:journals/corr/abs-2101-09995} pushes to recognize algorithmic fairness from a more inclusive lens.
    
    Bias crops up in multiple steps of the pipeline \cite{https://doi.org/10.1111/lnc3.12432} \cite{sap-etal-2022-annotators}, including the annotation process, the training data, the input representations, model architecture, and the structure of the research design. Thus, measures to mitigate bias in one of these components alone will likely not suffice as a corrective measure, necessitating human intervention at different stages of the pipeline.
    
      Most work that addresses fairness in NLP addresses it from an Anglo-centric context, with comparatively significantly less work done in grammatically-gendered and low-resource languages. Their inability to capture social and cultural nuances and demographic variations is well-documented \cite{talat-etal-2022-reap}. Despite this, they are ubiquitous, with applications ranging diverse fields, from legal contexts to healthcare. That said, there is insufficient documentation of the harms that could stem from unfair models trained for downstream tasks involving natural language generation, despite \citet{Arnold2018SentimentBI, bhat-etal-2021-people, Buschek2021TheIO} indicating the influence of these systems on users. Apart from this, these NLP systems also reinforce and reproduce the social and racial hierarchies observed in society and fail to recognize underrepresented communities that are already marginalized \cite{dev-etal-2021-harms, https://doi.org/10.48550/arxiv.2202.11923}. The ramifications of neglecting these issues are diverse and far-reaching, from minor inconveniences for users in less harmful contexts to compromising their privacy as well as depriving them of opportunities and resources \cite{Cirillo2020, Kochling2020}.
      
    Finally, while the interplay and tradeoff between privacy, efficiency, and fairness in tabular data has received extensive examination \cite{https://doi.org/10.48550/arxiv.2010.03058,https://doi.org/10.48550/arxiv.2010.01285} comparatively fewer studies have been conducted in NLP \cite{tal-etal-2022-fewer, ahn-etal-2022-knowledge, https://doi.org/10.48550/arxiv.2211.04256}. 
    
    The contributions of this work center around drawing attention to the current state of research on fairness in the context of linguistic and cultural issues in non-English languages and in the context of multilingual models. While thorough survey studies such as \citet{sun-etal-2019-mitigating, DBLP:journals/corr/abs-2112-14168, bhatt-etal-2022-contextualizing} yield valuable insights into some of these aspects, none address the current state of the work in multilingual fairness. Our paper provides insights into the following:
    \begin{itemize}
        \vspace{-0.2cm}
        \item This work surveys and presents challenges and unanswered questions with respect to fairness in both monolingual and multilingual NLP.
        \vspace{-0.2cm}
        \item We analyze bias from both a linguistic and cultural lens for non-English languages and present a comprehensive overview of the literature in bias pertaining to grammatically gendered languages and multilinguality.
        \vspace{-0.2cm}
        \item We bring to the forefront challenges in multilingual fairness and begin a dialogue for creating more equitable systems for multilingual NLP.
    \end{itemize}

\section{Bias in Monolingual Setups for English}

\subsection{Metrics for Measurement}
Prior to delving into the complexities of fairness in multilingual systems, it is essential to first examine the prevalent biases and challenges in monolingual systems. By prefacing the discussion on bias in multilingual systems with an overview of the current state of fairness evaluation and identifying areas for improvement, we aim to shed light on the potential for similar issues to arise in multilingual systems, as many of the biases present in monolingual systems are likely to persist in multilingual contexts. Some of the initial work on analyzing biases in NLP models \cite{DBLP:journals/corr/BolukbasiCZSK16a} propose quantitative measures of evaluating bias in word embeddings. Broadly speaking, bias measures are subcategorized into \textbf{\textit{i) intrinsic}} and \textbf{\textit{ii) extrinsic}} measures. Intrinsic metrics quantify bias in the model's pre-trained representations, whereas extrinsic metrics deal with bias observed in the outputs of the downstream task the model is trained for.

\citet{doi:10.1126/science.aal4230, may-etal-2019-measuring,nadeem-etal-2021-stereoset, https://doi.org/10.48550/arxiv.2010.00133} are commonly used in papers evaluating language models for fairness. \citet{doi:10.1126/science.aal4230} proposes the Word Embedding Association Test (WEAT). A fundamental criticism of WEAT is that it can be exploited to overestimate the bias in a model \cite{ethayarajh-etal-2019-understanding}. The Sentence Encoder Association Test (SEAT) metric \cite{may-etal-2019-measuring} was proposed to address WEAT's limitation of measuring bias only over static word embeddings. SEAT is an adaptation of WEAT that allows us to measure bias over contextualized embeddings.

StereoSet \cite{nadeem-etal-2021-stereoset}, and CrowS-Pair \cite{https://doi.org/10.48550/arxiv.2010.00133} are crowdsourced datasets specifically geared toward measuring the model's stereotypical proclivity over multiple dimensions, which are inclusive of gender, race, and religion, among others. \citet{blodgett-etal-2021-stereotyping} points out the flaws in the data quality, such as invalid stereotype/anti-stereotype pairs, reliance on indirect group identifiers as a proxy for demographic identification, and logical incongruities in the sentence pairs.

Several other intrinsic measures and adaptations of the aforementioned ones have also been proposed \cite{kurita-etal-2019-measuring, DBLP:journals/corr/abs-2010-06032, https://doi.org/10.48550/arxiv.2104.07496, DBLP:journals/corr/abs-2109-03646}. Recent studies \cite{delobelle-etal-2022-measuring, meade-etal-2022-empirical} that perform comparative evaluations across these measures provide valuable insights into how and where the metrics can be used, along with their potential drawbacks.

\subsection{Intrinsic vs Extrinsic Evaluation}
While intrinsic measures are valuable in that they indicate the existence of representational bias in systems, the current literature on fairness evaluation largely concentrates on intrinsic metrics alone. Considerably less work has been done on addressing bias in extrinsic evaluation, with several downstream tasks needing concrete metrics to evaluate bias in their outputs. This is a pressing issue due to the lack of correlation between intrinsic and extrinsic measures \cite{https://doi.org/10.48550/arxiv.2012.15859, cao-etal-2022-intrinsic, delobelle-etal-2022-measuring}. As emphasized in \citet{orgad-belinkov-2022-choose}, incorporating extrinsic evaluation measures is crucial for several reasons, including the greater relevance of these metrics to bias mitigation objectives. Aside from this, evaluating fairness on the downstream task's outputs allows us to gauge more precisely how a particular demographic may be affected by the biases in the system.

Although work done in fairness evaluation in NLP primarily concentrates on monolingual studies, there remain several unanswered questions and inconclusive results. For instance, although \citet{may-etal-2019-measuring} claims to use semantically bleached templates, experiments in \citet{delobelle-etal-2022-measuring} suggest that they retain some degree of semantic significance. While several bias evaluation methods use template-based data, recent findings \cite{alnegheimish-etal-2022-using} suggest that this approach may be unreliable and advocate the use of natural sentence prompts.

\subsection{Fairness From the Lens of Multiple Social Dimensions}

The focus of much of the existing body of literature is on gender bias, with little that covers other dimensions like race and religion. Evaluation metrics should be able to evaluate harms in language models over the intersectionality of multiple identities, akin to what would realistically be expected in real-world data. While previous research \cite{talat-etal-2022-reap, DBLP:journals/corr/abs-2102-04130} has emphasized the importance of fairness evaluation and mitigation over intersectional identities, there is relatively sparse work that attempts to address the same \cite{DBLP:journals/corr/abs-1911-01485, https://doi.org/10.48550/arxiv.2109.10441, https://doi.org/10.48550/arxiv.2110.00521, lalor-etal-2022-benchmarking, camara-etal-2022-mapping}. It is also crucial to gauge if reducing bias across one dimension could affect biases in the other dimensions. Most fairness measures do not account for the intersectionality of identities and standards of justice outside the predominantly Western sphere of distributive justice \cite{DBLP:journals/corr/abs-2101-09995, Lundgard_2020}.

Whilst there has been an increase in proposing novel methods to mitigate bias in language models, there needs to be more work in benchmarking these debiasing techniques to assess their relative effectiveness. \citet{meade-etal-2022-empirical} represents a step forward in this direction. Despite criticism \cite{ethayarajh-etal-2019-understanding, blodgett-etal-2021-stereotyping} of some evaluation metrics, they are still consistently used (and not always in conjunction with other metrics) in bias evaluation studies.

\section{Linguistic Aspects}\label{LinguisticAspects}

The linguistic variations between languages pose additional problems in the realm of multilingual NLP. Take, for example, the concept of gender, which has multiple definitions in linguistic terms (namely, \textbf{grammatical, referential, lexical and bio-social gender}) \cite{DBLP:journals/corr/abs-2112-14168}. Section \ref{grammaticallygendered} delves into how the grammatically gendered nature of languages can affect bias in multilingual and monolingual spaces alike. \textbf{Referential} gender, on the other hand, deals with terms that referentially address a person's gender, such as pronouns. Terms that non-referentially describe gender fall under the umbrella of \textbf{lexical} gender, and the \textbf{bio-social} definition of gender involves a mixture of phenotypic traits, gender expression, and identity as well as societal and cultural aspects that influence them \cite{Ackerman2019SyntacticAC}.

Although initial forays into this field investigate bias caused by grammatical gender, problems in these systems can also crop up due to the other definitions of gender. Referential gender terms are not always aligned when used in conjunction with lexically gendered terms, particularly with respect to pronoun-based anaphors for queer-identifying individuals. Several default assumptions regarding the individual's gender identity are made as a consequence \cite{cao-daume-iii-2021-toward}.

There are multiple varying forms of pronoun complexity \cite{inbook, doi:10.1086/465527}. Apart from this, there are instances of substantial variations in their linguistic forms even among languages within a specific region, as highlighted in \citet{nair2013tribal}. Linguistics also involves the presence of constructs like deictic pronouns and honorific pronouns \cite{Goddard2005TheLO}, which in some cases can lead to the pronouns used to reference someone changing based on their social dynamic within the community \cite{lauscher-etal-2022-welcome}. These linguistic aspects represent another line of work that must be addressed for lower-resourced communities that communicate using languages that utilize these.

Lexical gender, while non-referential, finds its own challenges due to the variation of these terms across languages. For example, while certain relationships with individuals in a family may have an exact mapping in other languages, more often than not (particularly with Southeast Asian languages), there is no precise mapping, and the system ends up making an approximation or ignoring the term altogether. Such issues may also be likely to perforate to other axes such as race, religion, caste, and so forth. In particular, considering that one method of training multilingual embeddings relies on alignment-based approaches, it is imperative that we keep in mind how these design choices could affect the representations of these terms.

Whilst utilizing linguistic features in methods to evaluate and mitigate gender bias is a relatively new field of study, previous work has demonstrated that additional linguistic context can result in performance gains \cite{volkova-etal-2013-exploring, wallace-etal-2014-humans}, thus in alignment with the claim from \newcite{hovy-yang-2021-importance} that LMs must utilize social context to be able to reach human-level performance on tasks. \newcite{sun-etal-2021-cross} utilizes linguistic features to capture cross-cultural similarities, and thus, to select languages that are optimal for cross-lingual transfer. However, it is essential to acknowledge that languages are susceptible to cultural and linguistic shifts that occur at both global and local levels over time, as noted in \citet{hamilton-etal-2016-cultural}. Pretrained models also have the capability to embed sociodemographic information, as evinced by \citet{lauscher-etal-2022-socioprobe}.

It has also been noted that other linguistic forms of gender do not translate well to sociological gender \cite{cao-daume-iii-2021-toward}. Furthermore, the scarcity of non-binary gender options in different languages can lead to the misgendering of non-binary individuals in these languages, as they may be constricted to fit into a binarized definition of sociological gender.

\subsection{Grammatically Gendered Languages} \label{grammaticallygendered}

Linguistics recognizes multiple forms of gender \cite{cao-daume-iii-2020-toward}, as observed in grammatically gendered languages where most or all nouns, including those referring to inanimate objects, possess a syntactic concept of gender. These languages can have anywhere between 2 to 20 forms of grammatical gender divisions. There has been an almost exclusive focus on English for evaluating gender bias, even in the setting of monolingual models and systems. English, however, is not a grammatically-gendered language. This may limit the transferability of techniques used for bias evaluation and mitigation to other languages that are grammatically gendered.

\citet{zhou-etal-2019-examining} examines bias from the view of grammatically gendered languages by decomposing the gendered information of words in the embedding space into two components; i) semantic and ii) syntactic. For instance, the Spanish word for "man" (\textit{hombre}) is both semantically and syntactically gendered. However, the Spanish word for "water" (\textit{agua}) is not semantically gendered but is considered a feminine noun. The proximity of female occupation words to the feminine side and male occupation words to the masculine side of the semantic gender direction suggests the presence of bias in these Spanish embeddings. \citet{zhou-etal-2019-examining} also demonstrates via experiments on bilingual embeddings that, post-alignment, masculine-gendered words are closer to the English equivalent of the occupation words than feminine-gendered ones. The paper also proposes bias mitigation methods and demonstrates that the quality of the embeddings is preserved via word-translation experiments. Nevertheless, the validity of these mitigation measures would need to be verified by testing them on downstream tasks. \citet{gonen-etal-2019-grammatical} show that grammatical gender affects the word representations in Italian and German and that inanimate nouns end up being closer to words of the same gender. They propose to address this through the precise use of a language-specific morphological tool and a careful approach to removing all the gender signals from a given text.

The grammatical properties of a language might show some interesting properties to be taken into account when dealing with the fairness of large language models, particularly for gender bias. Studies directed toward them could yield insights into observable trends across language families, with \citet{gonen-etal-2019-grammatical} demonstrating how the alignment of languages in the embedding space is negatively affected by grammatical gender. They could also prove helpful when analyzing bias in multilingual models, where both grammatically gendered and non-gendered languages are aligned to the same embedding space. The research and datasets available for extrinsic evaluation over other languages remain an area with scope for improvement.

Apart from these grammatical properties that affect the results we observe, the translation of existing bias evaluation datasets into other languages to create parallel corpora does not suffice when dealing with languages apart from English. This is partly because most languages are inherently rooted in cultural context. Any data curated for these languages must incorporate socio-cultural and linguistic aspects unique to the language/region. Depriving NLP systems of cultural context could consequently lead to entire axes over which social biases are measured being ignored. The cultural significance of words and phrases in various languages can vary significantly, as demonstrated in \citet{https://doi.org/10.48550/arxiv.2211.10780}, as well as in characteristics such as metaphorical tendencies \cite{gutierrez-etal-2016-detecting} and communication styles \cite{miehle-etal-2016-cultural, SUSZCZYNSKA19991053}. \newcite{hovy-yang-2021-importance} includes an overview and critique of this in the current state of NLP literature, which they claim adopts an oversimplified view and focuses on the information content alone while ignoring the social context of this content. \newcite{Milios2022AnAO, espana-bonet-etal-2022-attenuation} illustrate the inefficiency of direct translation methods, and \newcite{espana-bonet-etal-2022-attenuation} advocates for the creation of culturally-sensitive datasets for fairness assessment. However, \newcite{kaneko-etal-2022-gender} proposes a way to generate parallel corpora for other languages that bears high correlation with human bias annotations.

\section{Multilingual Models}

Multilingual spaces allow the embeddings of multiple languages to be aligned so that the mappings of every word to its equivalent in other languages are close to each in these embedding spaces. There are numerous ways of training multilingual language models \cite{hedderich-etal-2021-survey} using monolingual and unlabeled data. Multilingual language models can improve cross-lingual performance on low-resource languages leveraging the data available to higher-resourced languages up to a certain number of languages. Beyond a point, however, the performance across these languages on cross-lingual and monolingual tasks begins to dip as the number of languages increases \cite{conneau-etal-2020-unsupervised}. However, few studies explore the impact of multilingual training on biases. \citet{hovy-yang-2021-importance} illustrate how language and culture share a strong association, and \newcite{khani-etal-2021-cultural, sun-etal-2021-cross} reveal that geographical and cultural proximity among languages could enhance the performance of models. 

Languages provide much insight into a society's cultural norms, ideologies, and belief systems \cite{hershcovich-etal-2022-challenges, wilson-etal-2016-disentangling}. Often, the properties unique to a language are not clearly mapped to other languages or even other dialects within a language, with no direct translations available for several phrases and terminology. Whether or not language models can retain this cultural information and context while utilizing information from higher-resourced languages still requires investigation.

\subsection{An Outline of Fairness Evaluation in the Context of Multilinguality}

Several datasets have been put forward for the purpose of multilingual evaluation, and Table \ref{tab:datasets} describes these datasets along with details regarding their utility. These include the languages they cover, whether or not they evaluate bias over pretrained representations or a downstream task, and the downstream tasks and dimensions they cater toward.

\begin{table*}[h]
\begin{adjustbox}{max width=1\textwidth}
\begin{tabular}{|l|l|l|l|l|}
\hline
\textbf{Dataset}                                                & \textbf{Languages}                                                                                                     & \multicolumn{1}{c|}{\textbf{Task}} & \textbf{Metric}     & \multicolumn{1}{c|}{\textbf{Dimensions}}                                             \\ \hline
{\bf \href{https://github.com/MSR-LIT/MultilingualBias}{\color{darkblue}\citet{zhao-etal-2020-gender}}}                     & English, Spanish, German, French                                                                                       & Text Classification                & I, E & Gender                                                                               \\ \hline
{\bf \href{https://github.com/xiaoleihuang/DomainFairness}{\color{darkblue}\citet{huang-2022-easy}}}
                  & English, Italian, Portuguese, Spanish                                                                                  & Text Classification                & E           & Gender                                                                               \\ \hline
{\bf \href{https://github.com/kanekomasahiro/bias\_eval\_in\_multiple\_mlm}{\color{darkblue}\citet{kaneko-etal-2022-gender}}}
 & \begin{tabular}[c]{@{}l@{}}German, Japanese, Arabic, Spanish, \\ Portuguese, Russian, Indonesian, Chinese\end{tabular} & Masked Language Modelling          & I           & Gender                                                                               \\ \hline
{\bf \href{https://github.com/ascamara/ml-intersectionality}{\color{darkblue}\citet{camara-etal-2022-mapping}}}
& English, Arabic, Spanish                                                                                               & Text Classification                & E           & \begin{tabular}[c]{@{}l@{}}Gender, Race/Ethnicity,\\ Intersection\end{tabular}       \\ \hline
{\bf \href{https://github.com/liangsheng02/densray-debiasing/}{\color{darkblue}\citet{liang-etal-2020-monolingual}}}
              & English, Chinese                                                                                                       & Masked Language Modelling          & I           & Gender                                                                               \\ \hline
{\bf \href{https://github.com/xiaoleihuang/Multilingual\_Fairness\_LREC}{\color{darkblue}\citet{huang-etal-2020-multilingual}}}
    & \begin{tabular}[c]{@{}l@{}}English, Italian, Portuguese, \\ Spanish, Polish\end{tabular}                               & Text Classification                & E           & \begin{tabular}[c]{@{}l@{}}Age, Country, Gender, \\ Race/Ethnicity\end{tabular}      \\ \hline
{\bf \href{https://github.com/coastalcph/fairlex}{\color{darkblue}\citet{chalkidis-etal-2022-fairlex}}}                           & \begin{tabular}[c]{@{}l@{}}English, German, French,\\ Italian and Chinese\end{tabular}                                 & Text Classification                & E           & \begin{tabular}[c]{@{}l@{}}Gender, Age, Region, \\ Language, Legal Area\end{tabular} \\ \hline
\end{tabular}
\end{adjustbox}
\caption{Datasets for fairness evaluation beyond English. I = Intrinsic, E = Extrinsic}
\label{tab:datasets}
\end{table*}

\citet{zhao-etal-2020-gender} was among the first papers to quantify biases in multilingual spaces and does so using both extrinsic and intrinsic evaluation techniques. Their findings indicate that some factors that influence bias in multilingual embeddings include the language's linguistic properties, the target language used for the alignment of the embeddings, and transfer learning on these embeddings induces bias. Additionally, there is the possibility that non-Germanic languages do not align well with Germanic ones, and further work would be required to derive conclusions as to how this affects fairness measurements. 

\citet{huang-etal-2020-multilingual} released the first multilingual Twitter corpus for hate speech detection, annotated with the author's demographic attributes (age, country, gender, race/ethnicity), which allows for fairness evaluation across hate speech classifiers. Through experiments, they prove that variations in language, which are highly correlated with demographic attributes \cite{preotiuc-pietro-ungar-2018-user, articlesocio}, can result in biased classifiers. However, there are some promising results from \citet{liang-etal-2020-monolingual}, which proposes a novel debiasing method using \citet{dufter-schutze-2019-analytical}. While the multilingual model is originally debiased over English, results show its effectiveness for zero-shot debiasing over Chinese.

\citet{camara-etal-2022-mapping} measures both unisectional and intersectional social biases over gender, race, and ethnicity in multilingual language models. This is particularly relevant, as in a practical setting, treating identities as composites of various demographic attributes is a necessity. \citet{kaneko-etal-2022-gender} measures gender bias in masked language models and proposes a method to use parallel corpora to evaluate bias in languages shown to have high correlations with human bias annotations. In cases where manually annotated data doesn't exist, this could prove helpful.

Although there has been research on fairness in multimodal contexts \cite{10.1145/3514094.3534136, https://doi.org/10.48550/arxiv.2212.11261}, in a first-of-its-kind study, \citet{wang-etal-2022-assessing} looks at fairness from a multilingual view in multimodal representations. Whilst they find that multimodal representations may be individually fair, i.e., similar text representations across languages translate to similar images, this concept of fairness does not extend across multiple groups.

\citet{talat-etal-2022-reap} expresses criticism over the primary data source for multilingual large language models being English, which they claim is reflective of cultural imperialism. They also advocate for these models to be used only for languages they have been trained for to retain the cultural context unique to a language. The multilingual datasets commonly used tend to be parallel corpora derived directly from English translations, neglecting the socio-cultural nuances specific to a given language, as evidenced by the CommonCrawl corpora \cite{dodge-etal-2021-documenting}.

Moreover, recent literature \cite{al-kuwatly-etal-2020-identifying, https://doi.org/10.48550/arxiv.2205.00415, sap-etal-2022-annotators} presents us with yet another potential issue; lack of demographic variation in the annotation of these dataset results could contribute to bias in the pipeline. As of yet, several languages \cite{aji-etal-2022-one, joshi-etal-2020-state} (such as Hindi, Arabic, and Indonesian, which have tens to hundreds of million of native speakers) have had little to no fairness benchmarking datasets developed for them, an indicator that much remains to be done to develop more equitable language models.

\subsection{An Outline of Fairness Mitigation in the Context of Multilinguality}

Due to multilingual spaces being a composite of the embeddings of various languages with different linguistic and semantic properties, it would serve mitigation techniques well to consider these differences. Other methods could use these distinctions to reduce bias in downstream tasks. \citet{zhao-etal-2020-gender}, for one, show that balancing the corpus and transferring it to a grammatically gendered language's embedding space could reduce bias, and that using debiased embeddings could also aid with bias mitigation.

\citet{huang-2022-easy} takes inspiration from the FEDA domain adaption technique \cite{daume-iii-2007-frustratingly} to use it to mitigate bias in multilingual text classification and compares this with other mitigation methods. These debiasing baselines involve adversarial training,  masking out tokens associated with demographic groups, and instance weighting to reduce the impact of data instances that could lead to more biased classifiers. While \citet{liang-etal-2020-monolingual} show that zero-shot debiasing can be beneficial for this purpose, further study would be required to ascertain if this is a feasible possibility. 

\subsection{Problems in Multilingual Evaluation and Mitigation}

A major challenge in multilingual fairness is the lack of datasets (including parallel corpora) and literature for evaluation across tasks. Much of the research conducted in monolingual contexts has yet to be replicated in a multilingual setting, which would enable us to determine whether or not bias trends in monolingual spaces are directly transferable to multilingual contexts. Research and data resources also tend to neglect less-represented demographics, notably those local to a particular region. Further, datasets require thorough documentation, as variations in annotator information can result in different types of biases infiltrating the pipeline \cite{https://doi.org/10.48550/arxiv.2211.10780, joshi-etal-2016-cultural, bracewell-tomlinson-2012-language}. These could include attitudes towards other cultures and languages, which must be assessed and reported during data collection. Multilingual users speak multiple languages, and there is no work on evaluating bias in language contact settings such as code-switching. Certain axes along which systems may discriminate may be contained to a given region. Due to the underrepresented nature of marginalized identities (such as immigrant communities), models will likely not learn useful representations of these identities. 

\section{Culture}

Language and culture are intrinsically linked with each other. However, NLP research has historically placed a considerable emphasis on the information content of the data, as opposed to the contextual information surrounding the same data. \citet{hovy-yang-2021-importance} propose a broad taxonomy of 7 social factors that encompasses various aspects of this contextual information. This could be incorporated into models to improve performance and make them aware from a socio-cultural perspective.

The differences between a pair of languages or even a pair of dialects could reflect across multiple attributes; this could lead to variations in language's \textit{phonology, tone, text, and lexical forms}. Some of these attributes are controlled by the speaker and receiver involved. Despite evidence of gains in performance by leveraging these features, systems still retain the potential to discriminate against marginalized communities, as evinced in \citet{sap-etal-2019-risk}. This necessitates the proposal of evaluation methods to analyze the potential harms that people from different cultural backgrounds might expose themselves to via the use of such systems.

Multilingualism also entails the need to navigate the nuances of language, including the potential for stereotypes and discriminatory language, which may not have precise equivalents in other languages. Cultural taboos and stereotypes can be highly localized. As an example, pregnant or lactating women are discouraged from consuming nutritious food in certain cultures \cite{Meyer-Rochow2009}. Such contextual information might be underrepresented or nonexistent in the data that the model is exposed to. While some culture-specific behaviors may be prohibited or frowned upon in some parts of the world, there are yet other places that may encourage or remain indifferent to these very same behaviors.

Additionally, the axes we consider require to be treated differently in different cultural and linguistic settings. Take, for instance, gender. While gender has, for the most part, been treated as a binary variable in these studies, this does not echo what is observed in real-world settings, where several individuals have non-binary gender identities \cite{10.1145/3531146.3534627}. Non-binary gender identities encompass a broad spectrum of gender identities, and the term is generally considered an umbrella term for any identity outside the binary. The inability of models to incorporate this additional information on gender has subsequently led to them developing meaningless representations of non-binary genders in text \cite{dev-etal-2021-harms}. This translates to the systematic erasure of their identities. \newcite{baumler-rudinger-2022-recognition} show that much remains to be done concerning addressing non-binary identities outside the Western context. For instance, several non-binary identities, such as the Aravanis and the Māhūs (local to India and Hawaii, respectively) are likely to have little to no meaningful coverage in the training data of the models. These identities can also have unanswered nuances in literature; for example, the Acaults of Myanmar do not consider transsexualism, transvestism, and homosexuality to be distinct categories. This is also applicable to languages such as Arabana-Wangkangurru, which make use of deictic pronouns (previously discussed in Section \ref{LinguisticAspects}) \cite{lauscher-etal-2022-welcome, Hercus1994AGO}.

Further, given that models are highly suscept to the kind of data they are trained on, it is unlikely that our models can recognize that certain forms of prejudice are more frequent in specific socio-cultural environments than others. The targets of this discrimination are also likely to vary from region to region, another nuance that models find difficult to account for. India and Nepal, for instance, are two countries that still suffer from the effects of the hierarchy of a historically caste-based society that (despite sharing similar roots) bear differences in terms of representation of the various castes and how they are referred to \cite{articlejodhka, jraoarticle}. It is important to note that the ability of a system to incorporate information from these social factors to mitigate biases is task-dependent. Downstream tasks like machine translation and dialogue/response generation may depend more on cues related to speaker and receiver characteristics from the taxonomy proposed in \citet{hovy-yang-2021-importance} than other tasks. Extrinsic metrics for machine translation focus primarily on the gender bias of the mappings of nouns and pronouns from one language to another \cite{cho-etal-2019-measuring}. On the other hand, more open-ended, subjective tasks like NLG are prone to encoding underlying biases and stereotypes across multiple axes and reproducing these in their outputs \cite{10.1145/3278721.3278777}.

It is critical to consider intersectionality in these studies, as every individual is a composite of multiple identities across multiple axes. When conducting inquiries into the biased nature of these systems, we encourage researchers to use metrics that treat fairness as an intersectional concept and keep in line with the recommendations as suggested in \citet{talat-etal-2022-reap, blodgett-etal-2020-language} to document the affected demographics. Testing the validity and reliability of bias measurement and debiasing metrics is essential to ensuring the effectiveness of proposed methods \cite{blodgett-etal-2020-language}, and it is crucial to report any limitations of the same.

\section{Moving Towards Inclusive Systems in All Languages}

The issue of fairness in multilingualism presents a number of challenges. Although current practitioners encourage making systems multicultural and developing systems to be used only for specific cultural contexts \cite{talat-etal-2022-reap}, we posit that this may not be a viable solution due to various practical considerations. The vast diversity of cultures and ethnicities across the world presents significant difficulties in terms of creating equitable multilingual systems. Even within languages such as English, several dialectal variants, both of the regional and social kind \cite{10.1162/COLI_a_00258}, still need to be accounted for. \newcite{https://doi.org/10.48550/arxiv.1707.00061} is an example of how this could further stigmatize oppressed communities. Language and various social aspects related to language are ever-evolving. Modeling aspects such as lexical variants and the syntactical difference between languages, elements like phonology, and speech inflections in spoken language could contribute to the complexity of these systems. 

Several countries have diverse concentrations of people from all regions of the world with unique backgrounds. The intricacies of the social interactions resulting from the population's diverse linguistic backgrounds and issues arising from language contact make the study of the fairness of multilingual systems that would be deployed to cater to these populations essential. It is not possible to make models agnostic to demographic attributes. Even with the omission of certain attributes, models can still exhibit bias based on factors such as linguistic variations in dialect, or the linguistic features employed, as demonstrated by \citet{hovy-sogaard-2015-tagging} who highlight the improved performance of NLP systems on texts written by older individuals. The data that large language models (LLMs) are trained on tends to be biased towards certain demographic strata \cite{Olteanu2019SocialDB}. Although curating more diverse datasets and following recommendations to mitigate bias in the data pipeline would be a step forward to mitigating this problem \cite{b-etal-2021-overview}, various resource constraints could hinder this or make it impractical.

Due to all these challenges and the ubiquity of language technologies that are used by large populations of non-English speaking users, addressing fairness and bias, taking into account diverse linguistic, socio-linguistic, and cultural factors, is of utmost importance. Interdisciplinary and multicultural teams are crucial to identifying, measuring, and mitigating harms caused by bias in multilingual models. Better evaluation benchmarks covering diverse linguistic phenomena and cultures will lead to better fairness evaluation. 

Regarding data collection, as discussed in Section \ref{grammaticallygendered}, it would be prudent to avoid directly translating datasets for training or evaluation in applications where fairness is critical. As we have shown in this survey, it is not enough to collect datasets in multiple languages for measuring and mitigating bias, although even these are lacking for most languages worldwide. Zero-shot techniques that ignore the cultural nuances of a language should be used with care in fairness-critical applications, as linguistically similar languages may have different cultural values and vice versa. Finally, multilingual models and systems need to incorporate shared value systems that take into account diverse cultures, although some cultural differences may still go unacknowledged.

\section*{Limitations}
Our work surveys fairness literature in languages other than English, including bias measurement and mitigation strategies. Although we call out the fact that bias in literature is studied from an Anglo-centric point of view, it is conceivable that we miss many diverse perspectives on linguistic and cultural aspects of bias in different languages and cultures of the world due to the relatively heterogeneous background (in terms of nationality, ethnicity and field of study) of the authors. There may also be other relevant work in the social science literature that we may have missed including in this survey.

\bibliography{anthology,custom}

\begin{thebibliography}{109}
\expandafter\ifx\csname natexlab\endcsname\relax\def\natexlab#1{#1}\fi

\bibitem[{Ackerman(2019)}]{Ackerman2019SyntacticAC}
Lauren Ackerman. 2019.
\newblock Syntactic and cognitive issues in investigating gendered coreference.
\newblock \emph{Glossa}, 4.

\bibitem[{Ahn et~al.(2022)Ahn, Lee, Kim, and Oh}]{ahn-etal-2022-knowledge}
Jaimeen Ahn, Hwaran Lee, Jinhwa Kim, and Alice Oh. 2022.
\newblock \href {https://doi.org/10.18653/v1/2022.gebnlp-1.27} {Why knowledge
  distillation amplifies gender bias and how to mitigate from the perspective
  of {D}istil{BERT}}.
\newblock In \emph{Proceedings of the 4th Workshop on Gender Bias in Natural
  Language Processing (GeBNLP)}, pages 266--272, Seattle, Washington.
  Association for Computational Linguistics.

\bibitem[{Aji et~al.(2022)Aji, Winata, Koto, Cahyawijaya, Romadhony, Mahendra,
  Kurniawan, Moeljadi, Prasojo, Baldwin, Lau, and Ruder}]{aji-etal-2022-one}
Alham~Fikri Aji, Genta~Indra Winata, Fajri Koto, Samuel Cahyawijaya, Ade
  Romadhony, Rahmad Mahendra, Kemal Kurniawan, David Moeljadi, Radityo~Eko
  Prasojo, Timothy Baldwin, Jey~Han Lau, and Sebastian Ruder. 2022.
\newblock \href {https://doi.org/10.18653/v1/2022.acl-long.500} {One country,
  700+ languages: {NLP} challenges for underrepresented languages and dialects
  in {I}ndonesia}.
\newblock In \emph{Proceedings of the 60th Annual Meeting of the Association
  for Computational Linguistics (Volume 1: Long Papers)}, pages 7226--7249,
  Dublin, Ireland. Association for Computational Linguistics.

\bibitem[{Al~Kuwatly et~al.(2020)Al~Kuwatly, Wich, and
  Groh}]{al-kuwatly-etal-2020-identifying}
Hala Al~Kuwatly, Maximilian Wich, and Georg Groh. 2020.
\newblock \href {https://doi.org/10.18653/v1/2020.alw-1.21} {Identifying and
  measuring annotator bias based on annotators{'} demographic characteristics}.
\newblock In \emph{Proceedings of the Fourth Workshop on Online Abuse and
  Harms}, pages 184--190, Online. Association for Computational Linguistics.

\bibitem[{Alnegheimish et~al.(2022)Alnegheimish, Guo, and
  Sun}]{alnegheimish-etal-2022-using}
Sarah Alnegheimish, Alicia Guo, and Yi~Sun. 2022.
\newblock \href {https://doi.org/10.18653/v1/2022.naacl-main.203} {Using
  natural sentence prompts for understanding biases in language models}.
\newblock In \emph{Proceedings of the 2022 Conference of the North American
  Chapter of the Association for Computational Linguistics: Human Language
  Technologies}, pages 2824--2830, Seattle, United States. Association for
  Computational Linguistics.

\bibitem[{Arnold et~al.(2018)Arnold, Chauncey, and
  Gajos}]{Arnold2018SentimentBI}
Kenneth~C. Arnold, Krysta Chauncey, and Krzysztof~Z Gajos. 2018.
\newblock Sentiment bias in predictive text recommendations results in biased
  writing.
\newblock \emph{Proceedings of the 44th Graphics Interface Conference}.

\bibitem[{B et~al.(2021)B, Chandrabose, and
  Chakravarthi}]{b-etal-2021-overview}
Senthil~Kumar B, Aravindan Chandrabose, and Bharathi~Raja Chakravarthi. 2021.
\newblock \href {https://aclanthology.org/2021.ltedi-1.5} {An overview of
  fairness in data {--} illuminating the bias in data pipeline}.
\newblock In \emph{Proceedings of the First Workshop on Language Technology for
  Equality, Diversity and Inclusion}, pages 34--45, Kyiv. Association for
  Computational Linguistics.

\bibitem[{Ballard(1978)}]{doi:10.1086/465527}
William~L. Ballard. 1978.
\newblock \href {https://doi.org/10.1086/465527} {More on yuchi pronouns}.
\newblock \emph{International Journal of American Linguistics}, 44(2):103--112.

\bibitem[{Baumler and Rudinger(2022)}]{baumler-rudinger-2022-recognition}
Connor Baumler and Rachel Rudinger. 2022.
\newblock \href {https://doi.org/10.18653/v1/2022.naacl-main.250} {Recognition
  of they/them as singular personal pronouns in coreference resolution}.
\newblock In \emph{Proceedings of the 2022 Conference of the North American
  Chapter of the Association for Computational Linguistics: Human Language
  Technologies}, pages 3426--3432, Seattle, United States. Association for
  Computational Linguistics.

\bibitem[{Bender et~al.(2021)Bender, Gebru, McMillan-Major, and
  Shmitchell}]{10.1145/3442188.3445922}
Emily~M. Bender, Timnit Gebru, Angelina McMillan-Major, and Shmargaret
  Shmitchell. 2021.
\newblock \href {https://doi.org/10.1145/3442188.3445922} {On the dangers of
  stochastic parrots: Can language models be too big?}
\newblock In \emph{Proceedings of the 2021 ACM Conference on Fairness,
  Accountability, and Transparency}, FAccT '21, page 610–623, New York, NY,
  USA. Association for Computing Machinery.

\bibitem[{Bhat et~al.(2021)Bhat, Agashe, and Joshi}]{bhat-etal-2021-people}
Advait Bhat, Saaket Agashe, and Anirudha Joshi. 2021.
\newblock \href {https://aclanthology.org/2021.hcinlp-1.18} {How do people
  interact with biased text prediction models while writing?}
\newblock In \emph{Proceedings of the First Workshop on Bridging
  Human{--}Computer Interaction and Natural Language Processing}, pages
  116--121, Online. Association for Computational Linguistics.

\bibitem[{Bhatt et~al.(2022)Bhatt, Dev, Talukdar, Dave, and
  Prabhakaran}]{bhatt-etal-2022-contextualizing}
Shaily Bhatt, Sunipa Dev, Partha Talukdar, Shachi Dave, and Vinodkumar
  Prabhakaran. 2022.
\newblock \href {https://aclanthology.org/2022.aacl-main.55}
  {Re-contextualizing fairness in {NLP}: The case of {I}ndia}.
\newblock In \emph{Proceedings of the 2nd Conference of the Asia-Pacific
  Chapter of the Association for Computational Linguistics and the 12th
  International Joint Conference on Natural Language Processing (Volume 1: Long
  Papers)}, pages 727--740, Online only. Association for Computational
  Linguistics.

\bibitem[{Blodgett et~al.(2020)Blodgett, Barocas, Daum{\'e}~III, and
  Wallach}]{blodgett-etal-2020-language}
Su~Lin Blodgett, Solon Barocas, Hal Daum{\'e}~III, and Hanna Wallach. 2020.
\newblock \href {https://doi.org/10.18653/v1/2020.acl-main.485} {Language
  (technology) is power: A critical survey of {``}bias{''} in {NLP}}.
\newblock In \emph{Proceedings of the 58th Annual Meeting of the Association
  for Computational Linguistics}, pages 5454--5476, Online. Association for
  Computational Linguistics.

\bibitem[{Blodgett et~al.(2021)Blodgett, Lopez, Olteanu, Sim, and
  Wallach}]{blodgett-etal-2021-stereotyping}
Su~Lin Blodgett, Gilsinia Lopez, Alexandra Olteanu, Robert Sim, and Hanna
  Wallach. 2021.
\newblock \href {https://doi.org/10.18653/v1/2021.acl-long.81} {Stereotyping
  {N}orwegian salmon: An inventory of pitfalls in fairness benchmark datasets}.
\newblock In \emph{Proceedings of the 59th Annual Meeting of the Association
  for Computational Linguistics and the 11th International Joint Conference on
  Natural Language Processing (Volume 1: Long Papers)}, pages 1004--1015,
  Online. Association for Computational Linguistics.

\bibitem[{Blodgett and
  O'Connor(2017)}]{https://doi.org/10.48550/arxiv.1707.00061}
Su~Lin Blodgett and Brendan O'Connor. 2017.
\newblock \href {https://doi.org/10.48550/ARXIV.1707.00061} {Racial disparity
  in natural language processing: A case study of social media african-american
  english}.

\bibitem[{Bolukbasi et~al.(2016)Bolukbasi, Chang, Zou, Saligrama, and
  Kalai}]{DBLP:journals/corr/BolukbasiCZSK16a}
Tolga Bolukbasi, Kai{-}Wei Chang, James~Y. Zou, Venkatesh Saligrama, and Adam
  Kalai. 2016.
\newblock \href {http://arxiv.org/abs/1607.06520} {Man is to computer
  programmer as woman is to homemaker? debiasing word embeddings}.
\newblock \emph{CoRR}, abs/1607.06520.

\bibitem[{Bracewell and Tomlinson(2012)}]{bracewell-tomlinson-2012-language}
David Bracewell and Marc Tomlinson. 2012.
\newblock \href {https://aclanthology.org/C12-2016} {The language of power and
  its cultural influence}.
\newblock In \emph{Proceedings of {COLING} 2012: Posters}, pages 155--164,
  Mumbai, India. The COLING 2012 Organizing Committee.

\bibitem[{Buschek et~al.(2021)Buschek, Zurn, and Eiband}]{Buschek2021TheIO}
Daniel Buschek, Martin Zurn, and Malin Eiband. 2021.
\newblock The impact of multiple parallel phrase suggestions on email input and
  composition behaviour of native and non-native english writers.
\newblock \emph{Proceedings of the 2021 CHI Conference on Human Factors in
  Computing Systems}.

\bibitem[{Caliskan et~al.(2017)Caliskan, Bryson, and
  Narayanan}]{doi:10.1126/science.aal4230}
Aylin Caliskan, Joanna~J. Bryson, and Arvind Narayanan. 2017.
\newblock \href {https://doi.org/10.1126/science.aal4230} {Semantics derived
  automatically from language corpora contain human-like biases}.
\newblock \emph{Science}, 356(6334):183--186.

\bibitem[{C{\^a}mara et~al.(2022)C{\^a}mara, Taneja, Azad, Allaway, and
  Zemel}]{camara-etal-2022-mapping}
Ant{\'o}nio C{\^a}mara, Nina Taneja, Tamjeed Azad, Emily Allaway, and Richard
  Zemel. 2022.
\newblock \href {https://doi.org/10.18653/v1/2022.ltedi-1.11} {Mapping the
  multilingual margins: Intersectional biases of sentiment analysis systems in
  {E}nglish, {S}panish, and {A}rabic}.
\newblock In \emph{Proceedings of the Second Workshop on Language Technology
  for Equality, Diversity and Inclusion}, pages 90--106, Dublin, Ireland.
  Association for Computational Linguistics.

\bibitem[{Cao et~al.(2022)Cao, Pruksachatkun, Chang, Gupta, Kumar, Dhamala, and
  Galstyan}]{cao-etal-2022-intrinsic}
Yang Cao, Yada Pruksachatkun, Kai-Wei Chang, Rahul Gupta, Varun Kumar, Jwala
  Dhamala, and Aram Galstyan. 2022.
\newblock \href {https://doi.org/10.18653/v1/2022.acl-short.62} {On the
  intrinsic and extrinsic fairness evaluation metrics for contextualized
  language representations}.
\newblock In \emph{Proceedings of the 60th Annual Meeting of the Association
  for Computational Linguistics (Volume 2: Short Papers)}, pages 561--570,
  Dublin, Ireland. Association for Computational Linguistics.

\bibitem[{Cao and Daum{\'e}~III(2020)}]{cao-daume-iii-2020-toward}
Yang~Trista Cao and Hal Daum{\'e}~III. 2020.
\newblock \href {https://doi.org/10.18653/v1/2020.acl-main.418} {Toward
  gender-inclusive coreference resolution}.
\newblock In \emph{Proceedings of the 58th Annual Meeting of the Association
  for Computational Linguistics}, pages 4568--4595, Online. Association for
  Computational Linguistics.

\bibitem[{Cao and Daum{\'e}~III(2021)}]{cao-daume-iii-2021-toward}
Yang~Trista Cao and Hal Daum{\'e}~III. 2021.
\newblock \href {https://doi.org/10.1162/coli_a_00413} {Toward gender-inclusive
  coreference resolution: An analysis of gender and bias throughout the machine
  learning lifecycle*}.
\newblock \emph{Computational Linguistics}, 47(3):615--661.

\bibitem[{Chalkidis et~al.(2022)Chalkidis, Pasini, Zhang, Tomada, Schwemer, and
  S{\o}gaard}]{chalkidis-etal-2022-fairlex}
Ilias Chalkidis, Tommaso Pasini, Sheng Zhang, Letizia Tomada, Sebastian
  Schwemer, and Anders S{\o}gaard. 2022.
\newblock \href {https://doi.org/10.18653/v1/2022.acl-long.301} {{F}air{L}ex: A
  multilingual benchmark for evaluating fairness in legal text processing}.
\newblock In \emph{Proceedings of the 60th Annual Meeting of the Association
  for Computational Linguistics (Volume 1: Long Papers)}, pages 4389--4406,
  Dublin, Ireland. Association for Computational Linguistics.

\bibitem[{Cho et~al.(2019)Cho, Kim, Kim, and Kim}]{cho-etal-2019-measuring}
Won~Ik Cho, Ji~Won Kim, Seok~Min Kim, and Nam~Soo Kim. 2019.
\newblock \href {https://doi.org/10.18653/v1/W19-3824} {On measuring gender
  bias in translation of gender-neutral pronouns}.
\newblock In \emph{Proceedings of the First Workshop on Gender Bias in Natural
  Language Processing}, pages 173--181, Florence, Italy. Association for
  Computational Linguistics.

\bibitem[{Choudhury and Deshpande(2021)}]{Choudhury_Deshpande_2021}
Monojit Choudhury and Amit Deshpande. 2021.
\newblock \href {https://doi.org/10.1609/aaai.v35i14.17505} {How linguistically
  fair are multilingual pre-trained language models?}
\newblock \emph{Proceedings of the AAAI Conference on Artificial Intelligence},
  35(14):12710--12718.

\bibitem[{Cirillo et~al.(2020)Cirillo, Catuara-Solarz, Morey, Guney, Subirats,
  Mellino, Gigante, Valencia, Rementeria, Chadha, and Mavridis}]{Cirillo2020}
Davide Cirillo, Silvina Catuara-Solarz, Czuee Morey, Emre Guney, Laia Subirats,
  Simona Mellino, Annalisa Gigante, Alfonso Valencia, Mar{\'i}a~Jos{\'e}
  Rementeria, Antonella~Santuccione Chadha, and Nikolaos Mavridis. 2020.
\newblock \href {https://doi.org/10.1038/s41746-020-0288-5} {Sex and gender
  differences and biases in artificial intelligence for biomedicine and
  healthcare}.
\newblock \emph{npj Digital Medicine}, 3(1):81.

\bibitem[{Conneau et~al.(2020)Conneau, Khandelwal, Goyal, Chaudhary, Wenzek,
  Guzm{\'a}n, Grave, Ott, Zettlemoyer, and
  Stoyanov}]{conneau-etal-2020-unsupervised}
Alexis Conneau, Kartikay Khandelwal, Naman Goyal, Vishrav Chaudhary, Guillaume
  Wenzek, Francisco Guzm{\'a}n, Edouard Grave, Myle Ott, Luke Zettlemoyer, and
  Veselin Stoyanov. 2020.
\newblock \href {https://doi.org/10.18653/v1/2020.acl-main.747} {Unsupervised
  cross-lingual representation learning at scale}.
\newblock In \emph{Proceedings of the 58th Annual Meeting of the Association
  for Computational Linguistics}, pages 8440--8451, Online. Association for
  Computational Linguistics.

\bibitem[{Daum{\'e}~III(2007)}]{daume-iii-2007-frustratingly}
Hal Daum{\'e}~III. 2007.
\newblock \href {https://aclanthology.org/P07-1033} {Frustratingly easy domain
  adaptation}.
\newblock In \emph{Proceedings of the 45th Annual Meeting of the Association of
  Computational Linguistics}, pages 256--263, Prague, Czech Republic.
  Association for Computational Linguistics.

\bibitem[{Delobelle et~al.(2022)Delobelle, Tokpo, Calders, and
  Berendt}]{delobelle-etal-2022-measuring}
Pieter Delobelle, Ewoenam Tokpo, Toon Calders, and Bettina Berendt. 2022.
\newblock \href {https://doi.org/10.18653/v1/2022.naacl-main.122} {Measuring
  fairness with biased rulers: A comparative study on bias metrics for
  pre-trained language models}.
\newblock In \emph{Proceedings of the 2022 Conference of the North American
  Chapter of the Association for Computational Linguistics: Human Language
  Technologies}, pages 1693--1706, Seattle, United States. Association for
  Computational Linguistics.

\bibitem[{Dev et~al.(2021)Dev, Monajatipoor, Ovalle, Subramonian, Phillips, and
  Chang}]{dev-etal-2021-harms}
Sunipa Dev, Masoud Monajatipoor, Anaelia Ovalle, Arjun Subramonian, Jeff
  Phillips, and Kai-Wei Chang. 2021.
\newblock \href {https://doi.org/10.18653/v1/2021.emnlp-main.150} {Harms of
  gender exclusivity and challenges in non-binary representation in language
  technologies}.
\newblock In \emph{Proceedings of the 2021 Conference on Empirical Methods in
  Natural Language Processing}, pages 1968--1994, Online and Punta Cana,
  Dominican Republic. Association for Computational Linguistics.

\bibitem[{Devinney et~al.(2022)Devinney, Bj\"{o}rklund, and
  Bj\"{o}rklund}]{10.1145/3531146.3534627}
Hannah Devinney, Jenny Bj\"{o}rklund, and Henrik Bj\"{o}rklund. 2022.
\newblock \href {https://doi.org/10.1145/3531146.3534627} {Theories of
  “gender” in nlp bias research}.
\newblock In \emph{2022 ACM Conference on Fairness, Accountability, and
  Transparency}, FAccT '22, page 2083–2102, New York, NY, USA. Association
  for Computing Machinery.

\bibitem[{Dodge et~al.(2021)Dodge, Sap, Marasovi{\'c}, Agnew, Ilharco,
  Groeneveld, Mitchell, and Gardner}]{dodge-etal-2021-documenting}
Jesse Dodge, Maarten Sap, Ana Marasovi{\'c}, William Agnew, Gabriel Ilharco,
  Dirk Groeneveld, Margaret Mitchell, and Matt Gardner. 2021.
\newblock \href {https://doi.org/10.18653/v1/2021.emnlp-main.98} {Documenting
  large webtext corpora: A case study on the colossal clean crawled corpus}.
\newblock In \emph{Proceedings of the 2021 Conference on Empirical Methods in
  Natural Language Processing}, pages 1286--1305, Online and Punta Cana,
  Dominican Republic. Association for Computational Linguistics.

\bibitem[{Dufter and Sch{\"u}tze(2019)}]{dufter-schutze-2019-analytical}
Philipp Dufter and Hinrich Sch{\"u}tze. 2019.
\newblock \href {https://doi.org/10.18653/v1/D19-1111} {Analytical methods for
  interpretable ultradense word embeddings}.
\newblock In \emph{Proceedings of the 2019 Conference on Empirical Methods in
  Natural Language Processing and the 9th International Joint Conference on
  Natural Language Processing (EMNLP-IJCNLP)}, pages 1185--1191, Hong Kong,
  China. Association for Computational Linguistics.

\bibitem[{Espa{\~n}a-Bonet and
  Barr\'on-Cede{\~n}o(2022)}]{espana-bonet-etal-2022-attenuation}
Cristina Espa{\~n}a-Bonet and Alberto Barr\'on-Cede{\~n}o. 2022.
\newblock \href {https://aclanthology.org/2022.emnlp-main.} {The (undesired)
  attenuation of human biases by multilinguality}.
\newblock In \emph{Proceedings of the 2022 Conference on Empirical Methods in
  Natural Language Processing}, pages~--, Online and Abu Dhabi, UAE.
  Association for Computational Linguistics.

\bibitem[{Ethayarajh et~al.(2019)Ethayarajh, Duvenaud, and
  Hirst}]{ethayarajh-etal-2019-understanding}
Kawin Ethayarajh, David Duvenaud, and Graeme Hirst. 2019.
\newblock \href {https://doi.org/10.18653/v1/P19-1166} {Understanding
  undesirable word embedding associations}.
\newblock In \emph{Proceedings of the 57th Annual Meeting of the Association
  for Computational Linguistics}, pages 1696--1705, Florence, Italy.
  Association for Computational Linguistics.

\bibitem[{Goddard(2005)}]{Goddard2005TheLO}
Cliff Goddard. 2005.
\newblock The languages of east and southeast asia: An introduction.

\bibitem[{Goldfarb-Tarrant et~al.(2020)Goldfarb-Tarrant, Marchant, Sanchez,
  Pandya, and Lopez}]{https://doi.org/10.48550/arxiv.2012.15859}
Seraphina Goldfarb-Tarrant, Rebecca Marchant, Ricardo~Muñoz Sanchez, Mugdha
  Pandya, and Adam Lopez. 2020.
\newblock \href {https://doi.org/10.48550/ARXIV.2012.15859} {Intrinsic bias
  metrics do not correlate with application bias}.

\bibitem[{Gonen et~al.(2019)Gonen, Kementchedjhieva, and
  Goldberg}]{gonen-etal-2019-grammatical}
Hila Gonen, Yova Kementchedjhieva, and Yoav Goldberg. 2019.
\newblock \href {https://doi.org/10.18653/v1/K19-1043} {How does grammatical
  gender affect noun representations in gender-marking languages?}
\newblock In \emph{Proceedings of the 23rd Conference on Computational Natural
  Language Learning (CoNLL)}, pages 463--471, Hong Kong, China. Association for
  Computational Linguistics.

\bibitem[{Guti{\'e}rrez et~al.(2016)Guti{\'e}rrez, Shutova, Lichtenstein,
  de~Melo, and Gilardi}]{gutierrez-etal-2016-detecting}
E.D. Guti{\'e}rrez, Ekaterina Shutova, Patricia Lichtenstein, Gerard de~Melo,
  and Luca Gilardi. 2016.
\newblock \href {https://doi.org/10.1162/tacl_a_00082} {Detecting
  cross-cultural differences using a multilingual topic model}.
\newblock \emph{Transactions of the Association for Computational Linguistics},
  4:47--60.

\bibitem[{Hamilton et~al.(2016)Hamilton, Leskovec, and
  Jurafsky}]{hamilton-etal-2016-cultural}
William~L. Hamilton, Jure Leskovec, and Dan Jurafsky. 2016.
\newblock \href {https://doi.org/10.18653/v1/D16-1229} {Cultural shift or
  linguistic drift? comparing two computational measures of semantic change}.
\newblock In \emph{Proceedings of the 2016 Conference on Empirical Methods in
  Natural Language Processing}, pages 2116--2121, Austin, Texas. Association
  for Computational Linguistics.

\bibitem[{Hassan et~al.(2021)Hassan, Huenerfauth, and
  Alm}]{https://doi.org/10.48550/arxiv.2110.00521}
Saad Hassan, Matt Huenerfauth, and Cecilia~Ovesdotter Alm. 2021.
\newblock \href {https://doi.org/10.48550/ARXIV.2110.00521} {Unpacking the
  interdependent systems of discrimination: Ableist bias in nlp systems through
  an intersectional lens}.

\bibitem[{Hedderich et~al.(2021)Hedderich, Lange, Adel, Str{\"o}tgen, and
  Klakow}]{hedderich-etal-2021-survey}
Michael~A. Hedderich, Lukas Lange, Heike Adel, Jannik Str{\"o}tgen, and
  Dietrich Klakow. 2021.
\newblock \href {https://doi.org/10.18653/v1/2021.naacl-main.201} {A survey on
  recent approaches for natural language processing in low-resource scenarios}.
\newblock In \emph{Proceedings of the 2021 Conference of the North American
  Chapter of the Association for Computational Linguistics: Human Language
  Technologies}, pages 2545--2568, Online. Association for Computational
  Linguistics.

\bibitem[{Henderson et~al.(2018)Henderson, Sinha, Angelard-Gontier, Ke, Fried,
  Lowe, and Pineau}]{10.1145/3278721.3278777}
Peter Henderson, Koustuv Sinha, Nicolas Angelard-Gontier, Nan~Rosemary Ke,
  Genevieve Fried, Ryan Lowe, and Joelle Pineau. 2018.
\newblock \href {https://doi.org/10.1145/3278721.3278777} {Ethical challenges
  in data-driven dialogue systems}.
\newblock In \emph{Proceedings of the 2018 AAAI/ACM Conference on AI, Ethics,
  and Society}, AIES '18, page 123–129, New York, NY, USA. Association for
  Computing Machinery.

\bibitem[{Hercus(1994)}]{Hercus1994AGO}
Luise Hercus. 1994.
\newblock A grammar of the arabana-wangkangurru language : Lake eyre basin,
  south australia.

\bibitem[{Hershcovich et~al.(2022)Hershcovich, Frank, Lent, de~Lhoneux, Abdou,
  Brandl, Bugliarello, Cabello~Piqueras, Chalkidis, Cui, Fierro, Margatina,
  Rust, and S{\o}gaard}]{hershcovich-etal-2022-challenges}
Daniel Hershcovich, Stella Frank, Heather Lent, Miryam de~Lhoneux, Mostafa
  Abdou, Stephanie Brandl, Emanuele Bugliarello, Laura Cabello~Piqueras, Ilias
  Chalkidis, Ruixiang Cui, Constanza Fierro, Katerina Margatina, Phillip Rust,
  and Anders S{\o}gaard. 2022.
\newblock \href {https://doi.org/10.18653/v1/2022.acl-long.482} {Challenges and
  strategies in cross-cultural {NLP}}.
\newblock In \emph{Proceedings of the 60th Annual Meeting of the Association
  for Computational Linguistics (Volume 1: Long Papers)}, pages 6997--7013,
  Dublin, Ireland. Association for Computational Linguistics.

\bibitem[{Hessenthaler et~al.(2022)Hessenthaler, Strubell, Hovy, and
  Lauscher}]{https://doi.org/10.48550/arxiv.2211.04256}
Marius Hessenthaler, Emma Strubell, Dirk Hovy, and Anne Lauscher. 2022.
\newblock \href {https://doi.org/10.48550/ARXIV.2211.04256} {Bridging fairness
  and environmental sustainability in natural language processing}.

\bibitem[{Hooker et~al.(2020)Hooker, Moorosi, Clark, Bengio, and
  Denton}]{https://doi.org/10.48550/arxiv.2010.03058}
Sara Hooker, Nyalleng Moorosi, Gregory Clark, Samy Bengio, and Emily Denton.
  2020.
\newblock \href {https://doi.org/10.48550/ARXIV.2010.03058} {Characterising
  bias in compressed models}.

\bibitem[{Hovy and Prabhumoye(2021)}]{https://doi.org/10.1111/lnc3.12432}
Dirk Hovy and Shrimai Prabhumoye. 2021.
\newblock \href {https://doi.org/https://doi.org/10.1111/lnc3.12432} {Five
  sources of bias in natural language processing}.
\newblock \emph{Language and Linguistics Compass}, 15(8):e12432.

\bibitem[{Hovy and S{\o}gaard(2015)}]{hovy-sogaard-2015-tagging}
Dirk Hovy and Anders S{\o}gaard. 2015.
\newblock \href {https://doi.org/10.3115/v1/P15-2079} {Tagging performance
  correlates with author age}.
\newblock In \emph{Proceedings of the 53rd Annual Meeting of the Association
  for Computational Linguistics and the 7th International Joint Conference on
  Natural Language Processing (Volume 2: Short Papers)}, pages 483--488,
  Beijing, China. Association for Computational Linguistics.

\bibitem[{Hovy and Yang(2021)}]{hovy-yang-2021-importance}
Dirk Hovy and Diyi Yang. 2021.
\newblock \href {https://doi.org/10.18653/v1/2021.naacl-main.49} {The
  importance of modeling social factors of language: Theory and practice}.
\newblock In \emph{Proceedings of the 2021 Conference of the North American
  Chapter of the Association for Computational Linguistics: Human Language
  Technologies}, pages 588--602, Online. Association for Computational
  Linguistics.

\bibitem[{Huang(2022)}]{huang-2022-easy}
Xiaolei Huang. 2022.
\newblock \href {https://doi.org/10.18653/v1/2022.naacl-main.52} {Easy
  adaptation to mitigate gender bias in multilingual text classification}.
\newblock In \emph{Proceedings of the 2022 Conference of the North American
  Chapter of the Association for Computational Linguistics: Human Language
  Technologies}, pages 717--723, Seattle, United States. Association for
  Computational Linguistics.

\bibitem[{Huang et~al.(2020)Huang, Xing, Dernoncourt, and
  Paul}]{huang-etal-2020-multilingual}
Xiaolei Huang, Linzi Xing, Franck Dernoncourt, and Michael~J. Paul. 2020.
\newblock \href {https://aclanthology.org/2020.lrec-1.180} {Multilingual
  {T}witter corpus and baselines for evaluating demographic bias in hate speech
  recognition}.
\newblock In \emph{Proceedings of the Twelfth Language Resources and Evaluation
  Conference}, pages 1440--1448, Marseille, France. European Language Resources
  Association.

\bibitem[{Jodhka and Shah(2010)}]{articlejodhka}
Surinder~S. Jodhka and Ghanshyam Shah. 2010.
\newblock Comparative contexts of discrimination: Caste and untouchability in
  south asia.
\newblock \emph{Economic and Political Weekly}, 45.

\bibitem[{Joshi et~al.(2016)Joshi, Bhattacharyya, Carman, Saraswati, and
  Shukla}]{joshi-etal-2016-cultural}
Aditya Joshi, Pushpak Bhattacharyya, Mark Carman, Jaya Saraswati, and Rajita
  Shukla. 2016.
\newblock \href {https://doi.org/10.18653/v1/W16-2111} {How do cultural
  differences impact the quality of sarcasm annotation?: A case study of
  {I}ndian annotators and {A}merican text}.
\newblock In \emph{Proceedings of the 10th {SIGHUM} Workshop on Language
  Technology for Cultural Heritage, Social Sciences, and Humanities}, pages
  95--99, Berlin, Germany. Association for Computational Linguistics.

\bibitem[{Joshi et~al.(2020)Joshi, Santy, Budhiraja, Bali, and
  Choudhury}]{joshi-etal-2020-state}
Pratik Joshi, Sebastin Santy, Amar Budhiraja, Kalika Bali, and Monojit
  Choudhury. 2020.
\newblock \href {https://doi.org/10.18653/v1/2020.acl-main.560} {The state and
  fate of linguistic diversity and inclusion in the {NLP} world}.
\newblock In \emph{Proceedings of the 58th Annual Meeting of the Association
  for Computational Linguistics}, pages 6282--6293, Online. Association for
  Computational Linguistics.

\bibitem[{Kaneko and
  Bollegala(2021)}]{https://doi.org/10.48550/arxiv.2104.07496}
Masahiro Kaneko and Danushka Bollegala. 2021.
\newblock \href {https://doi.org/10.48550/ARXIV.2104.07496} {Unmasking the mask
  -- evaluating social biases in masked language models}.

\bibitem[{Kaneko et~al.(2022)Kaneko, Imankulova, Bollegala, and
  Okazaki}]{kaneko-etal-2022-gender}
Masahiro Kaneko, Aizhan Imankulova, Danushka Bollegala, and Naoaki Okazaki.
  2022.
\newblock \href {https://doi.org/10.18653/v1/2022.naacl-main.197} {Gender bias
  in masked language models for multiple languages}.
\newblock In \emph{Proceedings of the 2022 Conference of the North American
  Chapter of the Association for Computational Linguistics: Human Language
  Technologies}, pages 2740--2750, Seattle, United States. Association for
  Computational Linguistics.

\bibitem[{Khani et~al.(2021)Khani, Tourni, Rasooli, Callison-Burch, and
  Wijaya}]{khani-etal-2021-cultural}
Nikzad Khani, Isidora Tourni, Mohammad~Sadegh Rasooli, Chris Callison-Burch,
  and Derry~Tanti Wijaya. 2021.
\newblock \href {https://doi.org/10.18653/v1/2021.naacl-main.19} {Cultural and
  geographical influences on image translatability of words across languages}.
\newblock In \emph{Proceedings of the 2021 Conference of the North American
  Chapter of the Association for Computational Linguistics: Human Language
  Technologies}, pages 198--209, Online. Association for Computational
  Linguistics.

\bibitem[{Kirk et~al.(2021)Kirk, Jun, Iqbal, Benussi, Volpin, Dreyer,
  Shtedritski, and Asano}]{DBLP:journals/corr/abs-2102-04130}
Hannah Kirk, Yennie Jun, Haider Iqbal, Elias Benussi, Filippo Volpin,
  Fr{\'{e}}d{\'{e}}ric~A. Dreyer, Aleksandar Shtedritski, and Yuki~Markus
  Asano. 2021.
\newblock \href {http://arxiv.org/abs/2102.04130} {How true is gpt-2? an
  empirical analysis of intersectional occupational biases}.
\newblock \emph{CoRR}, abs/2102.04130.

\bibitem[{K{\"o}chling and Wehner(2020)}]{Kochling2020}
Alina K{\"o}chling and Marius~Claus Wehner. 2020.
\newblock \href {https://doi.org/10.1007/s40685-020-00134-w} {Discriminated by
  an algorithm: a systematic review of discrimination and fairness by
  algorithmic decision-making in the context of hr recruitment and hr
  development}.
\newblock \emph{Business Research}, 13(3):795--848.

\bibitem[{Kumar et~al.()Kumar, Balachandran, Njoo, Anastasopoulos, and
  Tsvetkov}]{https://doi.org/10.48550/arxiv.2210.07700}
Sachin Kumar, Vidhisha Balachandran, Lucille Njoo, Antonios Anastasopoulos, and
  Yulia Tsvetkov.
\newblock \href {https://doi.org/10.48550/ARXIV.2210.07700} {Language
  generation models can cause harm: So what can we do about it? an actionable
  survey}.

\bibitem[{Kurita et~al.(2019)Kurita, Vyas, Pareek, Black, and
  Tsvetkov}]{kurita-etal-2019-measuring}
Keita Kurita, Nidhi Vyas, Ayush Pareek, Alan~W Black, and Yulia Tsvetkov. 2019.
\newblock \href {https://doi.org/10.18653/v1/W19-3823} {Measuring bias in
  contextualized word representations}.
\newblock In \emph{Proceedings of the First Workshop on Gender Bias in Natural
  Language Processing}, pages 166--172, Florence, Italy. Association for
  Computational Linguistics.

\bibitem[{Lalor et~al.(2022)Lalor, Yang, Smith, Forsgren, and
  Abbasi}]{lalor-etal-2022-benchmarking}
John Lalor, Yi~Yang, Kendall Smith, Nicole Forsgren, and Ahmed Abbasi. 2022.
\newblock \href {https://doi.org/10.18653/v1/2022.naacl-main.263} {Benchmarking
  intersectional biases in {NLP}}.
\newblock In \emph{Proceedings of the 2022 Conference of the North American
  Chapter of the Association for Computational Linguistics: Human Language
  Technologies}, pages 3598--3609, Seattle, United States. Association for
  Computational Linguistics.

\bibitem[{Lauscher et~al.(2022{\natexlab{a}})Lauscher, Bianchi, Bowman, and
  Hovy}]{lauscher-etal-2022-socioprobe}
Anne Lauscher, Federico Bianchi, Samuel~R. Bowman, and Dirk Hovy.
  2022{\natexlab{a}}.
\newblock \href {https://aclanthology.org/2022.emnlp-main.539} {{S}ocio{P}robe:
  What, when, and where language models learn about sociodemographics}.
\newblock In \emph{Proceedings of the 2022 Conference on Empirical Methods in
  Natural Language Processing}, pages 7901--7918, Abu Dhabi, United Arab
  Emirates. Association for Computational Linguistics.

\bibitem[{Lauscher et~al.(2022{\natexlab{b}})Lauscher, Crowley, and
  Hovy}]{https://doi.org/10.48550/arxiv.2202.11923}
Anne Lauscher, Archie Crowley, and Dirk Hovy. 2022{\natexlab{b}}.
\newblock \href {https://doi.org/10.48550/ARXIV.2202.11923} {Welcome to the
  modern world of pronouns: Identity-inclusive natural language processing
  beyond gender}.

\bibitem[{Lauscher et~al.(2022{\natexlab{c}})Lauscher, Crowley, and
  Hovy}]{lauscher-etal-2022-welcome}
Anne Lauscher, Archie Crowley, and Dirk Hovy. 2022{\natexlab{c}}.
\newblock \href {https://aclanthology.org/2022.coling-1.105} {Welcome to the
  modern world of pronouns: Identity-inclusive natural language processing
  beyond gender}.
\newblock In \emph{Proceedings of the 29th International Conference on
  Computational Linguistics}, pages 1221--1232, Gyeongju, Republic of Korea.
  International Committee on Computational Linguistics.

\bibitem[{Lauscher et~al.(2021)Lauscher, L{\"{u}}ken, and
  Glavas}]{DBLP:journals/corr/abs-2109-03646}
Anne Lauscher, Tobias L{\"{u}}ken, and Goran Glavas. 2021.
\newblock \href {http://arxiv.org/abs/2109.03646} {Sustainable modular
  debiasing of language models}.
\newblock \emph{CoRR}, abs/2109.03646.

\bibitem[{Liang et~al.(2020)Liang, Dufter, and
  Sch{\"u}tze}]{liang-etal-2020-monolingual}
Sheng Liang, Philipp Dufter, and Hinrich Sch{\"u}tze. 2020.
\newblock \href {https://doi.org/10.18653/v1/2020.coling-main.446} {Monolingual
  and multilingual reduction of gender bias in contextualized representations}.
\newblock In \emph{Proceedings of the 28th International Conference on
  Computational Linguistics}, pages 5082--5093, Barcelona, Spain (Online).
  International Committee on Computational Linguistics.

\bibitem[{Lindström(2008)}]{inbook}
Eva Lindström. 2008.
\newblock \href {https://doi.org/10.1075/slcs.94} {\emph{Language complexity
  and interlinguistic difficulty}}, page 217–242.

\bibitem[{Liu et~al.(2021)Liu, Bugliarello, Ponti, Reddy, Collier, and
  Elliott}]{liu-etal-2021-visually}
Fangyu Liu, Emanuele Bugliarello, Edoardo~Maria Ponti, Siva Reddy, Nigel
  Collier, and Desmond Elliott. 2021.
\newblock \href {https://doi.org/10.18653/v1/2021.emnlp-main.818} {Visually
  grounded reasoning across languages and cultures}.
\newblock In \emph{Proceedings of the 2021 Conference on Empirical Methods in
  Natural Language Processing}, pages 10467--10485, Online and Punta Cana,
  Dominican Republic. Association for Computational Linguistics.

\bibitem[{Lundgard(2020)}]{Lundgard_2020}
Alan Lundgard. 2020.
\newblock \href {https://doi.org/10.1145/3351095.3372838} {Measuring justice in
  machine learning}.
\newblock In \emph{Proceedings of the 2020 Conference on Fairness,
  Accountability, and Transparency}. {ACM}.

\bibitem[{Lyu et~al.(2020)Lyu, He, and
  Li}]{https://doi.org/10.48550/arxiv.2010.01285}
Lingjuan Lyu, Xuanli He, and Yitong Li. 2020.
\newblock \href {https://doi.org/10.48550/ARXIV.2010.01285} {Differentially
  private representation for nlp: Formal guarantee and an empirical study on
  privacy and fairness}.

\bibitem[{May et~al.(2019)May, Wang, Bordia, Bowman, and
  Rudinger}]{may-etal-2019-measuring}
Chandler May, Alex Wang, Shikha Bordia, Samuel~R. Bowman, and Rachel Rudinger.
  2019.
\newblock \href {https://doi.org/10.18653/v1/N19-1063} {On measuring social
  biases in sentence encoders}.
\newblock In \emph{Proceedings of the 2019 Conference of the North {A}merican
  Chapter of the Association for Computational Linguistics: Human Language
  Technologies, Volume 1 (Long and Short Papers)}, pages 622--628, Minneapolis,
  Minnesota. Association for Computational Linguistics.

\bibitem[{Meade et~al.(2022)Meade, Poole-Dayan, and
  Reddy}]{meade-etal-2022-empirical}
Nicholas Meade, Elinor Poole-Dayan, and Siva Reddy. 2022.
\newblock \href {https://doi.org/10.18653/v1/2022.acl-long.132} {An empirical
  survey of the effectiveness of debiasing techniques for pre-trained language
  models}.
\newblock In \emph{Proceedings of the 60th Annual Meeting of the Association
  for Computational Linguistics (Volume 1: Long Papers)}, pages 1878--1898,
  Dublin, Ireland. Association for Computational Linguistics.

\bibitem[{Meyer-Rochow(2009)}]{Meyer-Rochow2009}
Victor~Benno Meyer-Rochow. 2009.
\newblock \href {https://doi.org/10.1186/1746-4269-5-18} {Food taboos: their
  origins and purposes}.
\newblock \emph{Journal of Ethnobiology and Ethnomedicine}, 5(1):18.

\bibitem[{Miehle et~al.(2016)Miehle, Yoshino, Pragst, Ultes, Nakamura, and
  Minker}]{miehle-etal-2016-cultural}
Juliana Miehle, Koichiro Yoshino, Louisa Pragst, Stefan Ultes, Satoshi
  Nakamura, and Wolfgang Minker. 2016.
\newblock \href {https://doi.org/10.18653/v1/W16-3610} {Cultural communication
  idiosyncrasies in human-computer interaction}.
\newblock In \emph{Proceedings of the 17th Annual Meeting of the Special
  Interest Group on Discourse and Dialogue}, pages 74--79, Los Angeles.
  Association for Computational Linguistics.

\bibitem[{Milios and BehnamGhader(2022)}]{Milios2022AnAO}
Aristides Milios and Parishad BehnamGhader. 2022.
\newblock An analysis of social biases present in bert variants across multiple
  languages.
\newblock \emph{ArXiv}, abs/2211.14402.

\bibitem[{Mohamed et~al.(2022)Mohamed, Abdelfattah, Alhuwaider, Li, Zhang,
  Church, and Elhoseiny}]{https://doi.org/10.48550/arxiv.2211.10780}
Youssef Mohamed, Mohamed Abdelfattah, Shyma Alhuwaider, Feifan Li, Xiangliang
  Zhang, Kenneth~Ward Church, and Mohamed Elhoseiny. 2022.
\newblock \href {https://doi.org/10.48550/ARXIV.2211.10780} {Artelingo: A
  million emotion annotations of wikiart with emphasis on diversity over
  language and culture}.

\bibitem[{Nadeem et~al.(2021)Nadeem, Bethke, and
  Reddy}]{nadeem-etal-2021-stereoset}
Moin Nadeem, Anna Bethke, and Siva Reddy. 2021.
\newblock \href {https://doi.org/10.18653/v1/2021.acl-long.416} {{S}tereo{S}et:
  Measuring stereotypical bias in pretrained language models}.
\newblock In \emph{Proceedings of the 59th Annual Meeting of the Association
  for Computational Linguistics and the 11th International Joint Conference on
  Natural Language Processing (Volume 1: Long Papers)}, pages 5356--5371,
  Online. Association for Computational Linguistics.

\bibitem[{Nair(2013)}]{nair2013tribal}
Ravi Sankar~S Nair. 2013.
\newblock Tribal languages of kerala.

\bibitem[{Nangia et~al.(2020)Nangia, Vania, Bhalerao, and
  Bowman}]{https://doi.org/10.48550/arxiv.2010.00133}
Nikita Nangia, Clara Vania, Rasika Bhalerao, and Samuel~R. Bowman. 2020.
\newblock \href {https://doi.org/10.48550/ARXIV.2010.00133} {Crows-pairs: A
  challenge dataset for measuring social biases in masked language models}.

\bibitem[{Nguyen et~al.(2016)Nguyen, Doğruöz, Rosé, and
  de~Jong}]{10.1162/COLI_a_00258}
Dong Nguyen, A.~Seza Doğruöz, Carolyn~P. Rosé, and Franciska de~Jong. 2016.
\newblock \href {https://doi.org/10.1162/COLI_a_00258} {{Computational
  Sociolinguistics: A Survey}}.
\newblock \emph{Computational Linguistics}, 42(3):537--593.

\bibitem[{Olteanu et~al.(2019)Olteanu, Castillo, Diaz, and
  Kıcıman}]{Olteanu2019SocialDB}
Alexandra Olteanu, Carlos Castillo, Fernando~D. Diaz, and Emre Kıcıman. 2019.
\newblock Social data: Biases, methodological pitfalls, and ethical boundaries.
\newblock \emph{Frontiers in Big Data}, 2.

\bibitem[{Orgad and Belinkov(2022)}]{orgad-belinkov-2022-choose}
Hadas Orgad and Yonatan Belinkov. 2022.
\newblock \href {https://doi.org/10.18653/v1/2022.gebnlp-1.17} {Choose your
  lenses: Flaws in gender bias evaluation}.
\newblock In \emph{Proceedings of the 4th Workshop on Gender Bias in Natural
  Language Processing (GeBNLP)}, pages 151--167, Seattle, Washington.
  Association for Computational Linguistics.

\bibitem[{Osiapem(2007)}]{articlesocio}
Iyabo Osiapem. 2007.
\newblock \href {https://doi.org/10.1017/S0047404507070327} {Florian coulmas,
  sociolinguistics: The study of speakers' choices -}.
\newblock \emph{Language in Society - LANG SOC}, 36.

\bibitem[{Parmar et~al.(2022)Parmar, Mishra, Geva, and
  Baral}]{https://doi.org/10.48550/arxiv.2205.00415}
Mihir Parmar, Swaroop Mishra, Mor Geva, and Chitta Baral. 2022.
\newblock \href {https://doi.org/10.48550/ARXIV.2205.00415} {Don't blame the
  annotator: Bias already starts in the annotation instructions}.

\bibitem[{Preo{\c{t}}iuc-Pietro and
  Ungar(2018)}]{preotiuc-pietro-ungar-2018-user}
Daniel Preo{\c{t}}iuc-Pietro and Lyle Ungar. 2018.
\newblock \href {https://aclanthology.org/C18-1130} {User-level race and
  ethnicity predictors from {T}witter text}.
\newblock In \emph{Proceedings of the 27th International Conference on
  Computational Linguistics}, pages 1534--1545, Santa Fe, New Mexico, USA.
  Association for Computational Linguistics.

\bibitem[{Rao(2010)}]{jraoarticle}
Jasmine Rao. 2010.
\newblock The caste system: Effects on poverty in india, nepal and sri lanka.
\newblock \emph{Glob. Majority E-J.}, 1.

\bibitem[{Sambasivan et~al.(2021)Sambasivan, Arnesen, Hutchinson, Doshi, and
  Prabhakaran}]{DBLP:journals/corr/abs-2101-09995}
Nithya Sambasivan, Erin Arnesen, Ben Hutchinson, Tulsee Doshi, and Vinodkumar
  Prabhakaran. 2021.
\newblock \href {http://arxiv.org/abs/2101.09995} {Re-imagining algorithmic
  fairness in india and beyond}.
\newblock \emph{CoRR}, abs/2101.09995.

\bibitem[{Sap et~al.(2019)Sap, Card, Gabriel, Choi, and
  Smith}]{sap-etal-2019-risk}
Maarten Sap, Dallas Card, Saadia Gabriel, Yejin Choi, and Noah~A. Smith. 2019.
\newblock \href {https://doi.org/10.18653/v1/P19-1163} {The risk of racial bias
  in hate speech detection}.
\newblock In \emph{Proceedings of the 57th Annual Meeting of the Association
  for Computational Linguistics}, pages 1668--1678, Florence, Italy.
  Association for Computational Linguistics.

\bibitem[{Sap et~al.(2022)Sap, Swayamdipta, Vianna, Zhou, Choi, and
  Smith}]{sap-etal-2022-annotators}
Maarten Sap, Swabha Swayamdipta, Laura Vianna, Xuhui Zhou, Yejin Choi, and
  Noah~A. Smith. 2022.
\newblock \href {https://doi.org/10.18653/v1/2022.naacl-main.431} {Annotators
  with attitudes: How annotator beliefs and identities bias toxic language
  detection}.
\newblock In \emph{Proceedings of the 2022 Conference of the North American
  Chapter of the Association for Computational Linguistics: Human Language
  Technologies}, pages 5884--5906, Seattle, United States. Association for
  Computational Linguistics.

\bibitem[{Stanczak and Augenstein(2021)}]{DBLP:journals/corr/abs-2112-14168}
Karolina Stanczak and Isabelle Augenstein. 2021.
\newblock \href {http://arxiv.org/abs/2112.14168} {A survey on gender bias in
  natural language processing}.
\newblock \emph{CoRR}, abs/2112.14168.

\bibitem[{Subramanian et~al.(2021)Subramanian, Han, Baldwin, Cohn, and
  Frermann}]{https://doi.org/10.48550/arxiv.2109.10441}
Shivashankar Subramanian, Xudong Han, Timothy Baldwin, Trevor Cohn, and Lea
  Frermann. 2021.
\newblock \href {https://doi.org/10.48550/ARXIV.2109.10441} {Evaluating
  debiasing techniques for intersectional biases}.

\bibitem[{Sun et~al.(2021)Sun, Ahn, Park, Tsvetkov, and
  Mortensen}]{sun-etal-2021-cross}
Jimin Sun, Hwijeen Ahn, Chan~Young Park, Yulia Tsvetkov, and David~R.
  Mortensen. 2021.
\newblock \href {https://doi.org/10.18653/v1/2021.eacl-main.204}
  {Cross-cultural similarity features for cross-lingual transfer learning of
  pragmatically motivated tasks}.
\newblock In \emph{Proceedings of the 16th Conference of the European Chapter
  of the Association for Computational Linguistics: Main Volume}, pages
  2403--2414, Online. Association for Computational Linguistics.

\bibitem[{Sun et~al.(2019)Sun, Gaut, Tang, Huang, ElSherief, Zhao, Mirza,
  Belding, Chang, and Wang}]{sun-etal-2019-mitigating}
Tony Sun, Andrew Gaut, Shirlyn Tang, Yuxin Huang, Mai ElSherief, Jieyu Zhao,
  Diba Mirza, Elizabeth Belding, Kai-Wei Chang, and William~Yang Wang. 2019.
\newblock \href {https://doi.org/10.18653/v1/P19-1159} {Mitigating gender bias
  in natural language processing: Literature review}.
\newblock In \emph{Proceedings of the 57th Annual Meeting of the Association
  for Computational Linguistics}, pages 1630--1640, Florence, Italy.
  Association for Computational Linguistics.

\bibitem[{Suszczyńska(1999)}]{SUSZCZYNSKA19991053}
Małgorzata Suszczyńska. 1999.
\newblock \href {https://doi.org/https://doi.org/10.1016/S0378-2166(99)00047-8}
  {Apologizing in english, polish and hungarian: Different languages, different
  strategies}.
\newblock \emph{Journal of Pragmatics}, 31(8):1053--1065.

\bibitem[{Tal et~al.(2022)Tal, Magar, and Schwartz}]{tal-etal-2022-fewer}
Yarden Tal, Inbal Magar, and Roy Schwartz. 2022.
\newblock \href {https://doi.org/10.18653/v1/2022.gebnlp-1.13} {Fewer errors,
  but more stereotypes? the effect of model size on gender bias}.
\newblock In \emph{Proceedings of the 4th Workshop on Gender Bias in Natural
  Language Processing (GeBNLP)}, pages 112--120, Seattle, Washington.
  Association for Computational Linguistics.

\bibitem[{Talat et~al.(2022)Talat, N{\'e}v{\'e}ol, Biderman, Clinciu, Dey,
  Longpre, Luccioni, Masoud, Mitchell, Radev, Sharma, Subramonian, Tae, Tan,
  Tunuguntla, and Van Der~Wal}]{talat-etal-2022-reap}
Zeerak Talat, Aur{\'e}lie N{\'e}v{\'e}ol, Stella Biderman, Miruna Clinciu,
  Manan Dey, Shayne Longpre, Sasha Luccioni, Maraim Masoud, Margaret Mitchell,
  Dragomir Radev, Shanya Sharma, Arjun Subramonian, Jaesung Tae, Samson Tan,
  Deepak Tunuguntla, and Oskar Van Der~Wal. 2022.
\newblock \href {https://doi.org/10.18653/v1/2022.bigscience-1.3} {You reap
  what you sow: On the challenges of bias evaluation under multilingual
  settings}.
\newblock In \emph{Proceedings of BigScience Episode {\#}5 -- Workshop on
  Challenges {\&} Perspectives in Creating Large Language Models}, pages
  26--41, virtual+Dublin. Association for Computational Linguistics.

\bibitem[{Tan and Celis(2019)}]{DBLP:journals/corr/abs-1911-01485}
Yi~Chern Tan and L.~Elisa Celis. 2019.
\newblock \href {http://arxiv.org/abs/1911.01485} {Assessing social and
  intersectional biases in contextualized word representations}.
\newblock \emph{CoRR}, abs/1911.01485.

\bibitem[{Volkova et~al.(2013)Volkova, Wilson, and
  Yarowsky}]{volkova-etal-2013-exploring}
Svitlana Volkova, Theresa Wilson, and David Yarowsky. 2013.
\newblock \href {https://aclanthology.org/D13-1187} {Exploring demographic
  language variations to improve multilingual sentiment analysis in social
  media}.
\newblock In \emph{Proceedings of the 2013 Conference on Empirical Methods in
  Natural Language Processing}, pages 1815--1827, Seattle, Washington, USA.
  Association for Computational Linguistics.

\bibitem[{Wallace et~al.(2014)Wallace, Choe, Kertz, and
  Charniak}]{wallace-etal-2014-humans}
Byron~C. Wallace, Do~Kook Choe, Laura Kertz, and Eugene Charniak. 2014.
\newblock \href {https://doi.org/10.3115/v1/P14-2084} {Humans require context
  to infer ironic intent (so computers probably do, too)}.
\newblock In \emph{Proceedings of the 52nd Annual Meeting of the Association
  for Computational Linguistics (Volume 2: Short Papers)}, pages 512--516,
  Baltimore, Maryland. Association for Computational Linguistics.

\bibitem[{Wang et~al.(2022)Wang, Liu, and Wang}]{wang-etal-2022-assessing}
Jialu Wang, Yang Liu, and Xin Wang. 2022.
\newblock \href {https://doi.org/10.18653/v1/2022.findings-acl.211} {Assessing
  multilingual fairness in pre-trained multimodal representations}.
\newblock In \emph{Findings of the Association for Computational Linguistics:
  ACL 2022}, pages 2681--2695, Dublin, Ireland. Association for Computational
  Linguistics.

\bibitem[{Webster et~al.(2020)Webster, Wang, Tenney, Beutel, Pitler, Pavlick,
  Chen, and Petrov}]{DBLP:journals/corr/abs-2010-06032}
Kellie Webster, Xuezhi Wang, Ian Tenney, Alex Beutel, Emily Pitler, Ellie
  Pavlick, Jilin Chen, and Slav Petrov. 2020.
\newblock \href {http://arxiv.org/abs/2010.06032} {Measuring and reducing
  gendered correlations in pre-trained models}.
\newblock \emph{CoRR}, abs/2010.06032.

\bibitem[{Wilson et~al.(2016)Wilson, Mihalcea, Boyd, and
  Pennebaker}]{wilson-etal-2016-disentangling}
Steven Wilson, Rada Mihalcea, Ryan Boyd, and James Pennebaker. 2016.
\newblock \href {https://doi.org/10.18653/v1/W16-5619} {Disentangling topic
  models: A cross-cultural analysis of personal values through words}.
\newblock In \emph{Proceedings of the First Workshop on {NLP} and Computational
  Social Science}, pages 143--152, Austin, Texas. Association for Computational
  Linguistics.

\bibitem[{Wolfe and Caliskan(2022)}]{10.1145/3514094.3534136}
Robert Wolfe and Aylin Caliskan. 2022.
\newblock \href {https://doi.org/10.1145/3514094.3534136} {American == white in
  multimodal language-and-image ai}.
\newblock In \emph{Proceedings of the 2022 AAAI/ACM Conference on AI, Ethics,
  and Society}, AIES '22, page 800–812, New York, NY, USA. Association for
  Computing Machinery.

\bibitem[{Wolfe et~al.(2022)Wolfe, Yang, Howe, and
  Caliskan}]{https://doi.org/10.48550/arxiv.2212.11261}
Robert Wolfe, Yiwei Yang, Bill Howe, and Aylin Caliskan. 2022.
\newblock \href {https://doi.org/10.48550/ARXIV.2212.11261} {Contrastive
  language-vision ai models pretrained on web-scraped multimodal data exhibit
  sexual objectification bias}.

\bibitem[{Zhao et~al.(2020)Zhao, Mukherjee, Hosseini, Chang, and
  Hassan~Awadallah}]{zhao-etal-2020-gender}
Jieyu Zhao, Subhabrata Mukherjee, Saghar Hosseini, Kai-Wei Chang, and Ahmed
  Hassan~Awadallah. 2020.
\newblock \href {https://doi.org/10.18653/v1/2020.acl-main.260} {Gender bias in
  multilingual embeddings and cross-lingual transfer}.
\newblock In \emph{Proceedings of the 58th Annual Meeting of the Association
  for Computational Linguistics}, pages 2896--2907, Online. Association for
  Computational Linguistics.

\bibitem[{Zhou et~al.(2019)Zhou, Shi, Zhao, Huang, Chen, Cotterell, and
  Chang}]{zhou-etal-2019-examining}
Pei Zhou, Weijia Shi, Jieyu Zhao, Kuan-Hao Huang, Muhao Chen, Ryan Cotterell,
  and Kai-Wei Chang. 2019.
\newblock \href {https://doi.org/10.18653/v1/D19-1531} {Examining gender bias
  in languages with grammatical gender}.
\newblock In \emph{Proceedings of the 2019 Conference on Empirical Methods in
  Natural Language Processing and the 9th International Joint Conference on
  Natural Language Processing (EMNLP-IJCNLP)}, pages 5276--5284, Hong Kong,
  China. Association for Computational Linguistics.

\end{thebibliography}
\bibliographystyle{acl_natbib}

\end{document}